\documentclass[twoside,leqno,twocolumn]{article}  
\usepackage{ltexpprt}

\newif\ifapx
\apxfalse % uncomment if you dont want appendix

\newcommand{\ourmaintitle}{Causal Inference on Multivariate and Mixed-Type Data}
\newcommand{\ourtitle}{\ourmaintitle}
\newcommand{\ourmethode}{\ensuremath{\textsc{Crack}_{\delta}}\xspace}
\newcommand{\ourmethodo}{\ensuremath{\textsc{Crack}_{\Delta}}\xspace}
\newcommand{\ourmethod}{\textsc{Crack}\xspace}
\newcommand{\oururl}{\url{http://eda.mmci.uni-saarland.de/crack/}}
\newcommand{\codeurl}{\oururl}

\makeatletter
\newif\if@restonecol
\makeatother

\usepackage{pgfplots}
\usepgfplotslibrary{fillbetween}
\usepackage{latexsym}
\usepackage{subcaption} % use minipage instead!
\usepackage{balance}
\usepackage[ruled, vlined, nofillcomment, linesnumbered]{algorithm2e}
\usepackage{etex}
\usepackage{fixltx2e}
\usepackage[english]{babel}
\usepackage{eda} % The EDA package, includes many default packages

\usepackage[pdftitle={\ourtitle}, pdfauthor={Alexander Marx and Jilles Vreeken}, hidelinks, hyperfootnotes = false]{hyperref}

\usepackage{pifont}
\newcommand{\cmark}{\ding{51}}

\graphicspath{ {./figs/} }

\ifpdf
\usetikzlibrary{external}
\fi

\SetKwComment{tcpas}{\{}{\}}
\SetCommentSty{textnormal}
\SetArgSty{textnormal}
\SetKw{False}{false}
\SetKw{True}{true}
\SetKw{Null}{null}
\SetKwInOut{Output}{output}
\SetKwInOut{Input}{input}
\SetKw{AND}{and}
\SetKw{OR}{or}
\SetKw{Continue}{continue}

\newcommand{\origo}{\textsc{Origo}\xspace}
\newcommand{\ltr}{\textsc{LTR}\xspace}
\newcommand{\ktr}{\textsc{KTR}\xspace}
\newcommand{\ergo}{\textsc{Ergo}\xspace}

\newcommand{\igci}{\textsc{IGCI}\xspace}
\newcommand{\slope}{\textsc{Slope}\xspace}
\newcommand{\cure}{\textsc{Cure}\xspace}

\newcommand{\TrivialTree}{\textsc{TrivialTree}\xspace}
\newcommand{\SplitOrRegress}{\textsc{RefineLeaf}\xspace}
\newcommand{\RobustMinDiff}{\textsc{RobustMinDiff}\xspace}
\newcommand{\GreedyCrack}{\textsc{Crack}\xspace}

\newcommand{\dom}{\ensuremath{\mathbb{D}}}
\newcommand{\res}{\ensuremath{\mathit{res}}}
\newcommand{\internal}{\ensuremath{\mathit{int}}}

\newcommand{\node}{\ensuremath{\mathit{v}}}

\newcommand{\leaf}{\ensuremath{\mathit{l}}}
\newcommand{\lvs}{\ensuremath{\mathit{lvs}}}

\newcommand{\tree}{\ensuremath{T}}

\newcommand{\depgraph}{\ensuremath{\mathcal{G}}}
\newcommand{\model}{\ensuremath{M}}
\renewcommand{\models}{\ensuremath{\mathcal{M}}}

\newcommand{\attributes}{\ensuremath{A}}

\newcommand{\num}{\ensuremath{\mathit{num}}\xspace}
\newcommand{\nom}{\ensuremath{\mathit{nom}}\xspace}

\newcommand{\tsplit}{\ensuremath{\mathit{split}}\xspace}
\newcommand{\tregress}{\ensuremath{\mathit{reg}}\xspace}

\newcommand{\LN}{\ensuremath{L_\mathbb{N}}}

\newcommand{\NCI}{\ensuremath{\mathit{NCI}}\xspace}

\newcommand{\data}{\ensuremath{D}}
\newcommand{\dataX}{\ensuremath{X}}
\newcommand{\dataY}{\ensuremath{Y}}

\newcommand{\XtoY}{{X \rightarrow Y}\xspace}
\newcommand{\YtoX}{{Y \rightarrow X}\xspace}

\newcommand{\DXY}{\Delta_{\XtoY}\xspace}
\newcommand{\DYX}{\Delta_{\YtoX}\xspace}

\newcommand{\cause}{\ensuremath{\mathit{cause}}\xspace}
\newcommand{\effect}{\ensuremath{\mathit{effect}}\xspace}

	\tikzstyle{flatlabel}  = [above, font = \tiny, inner sep = 1pt, text = black]
	\tikzstyle{flatlabelb}  = [below, font = \tiny, inner sep = 1pt, text = black]
	\tikzstyle{slopelabel}  = [sloped, above, font = \tiny, inner sep = 1pt, text = black]
	\tikzstyle{slopelabelb}  = [sloped, below, font = \tiny, inner sep = 1pt, text = black]

\usepackage{tikz}
\usepackage{pgfplots}
\usepackage{pgfplotstable}
\usetikzlibrary{arrows,shapes,positioning,fit,calc,matrix,trees}
\usetikzlibrary{external}
\pgfdeclareplotmark{btriangle}{%
  \pgfpathmoveto{\pgfpoint{-\pgfplotmarksize}{\pgfplotmarksize}}
  \pgfpathlineto{\pgfpoint{0pt}{-\pgfplotmarksize}}
  \pgfpathlineto{\pgfpoint{\pgfplotmarksize}{\pgfplotmarksize}}
  \pgfpathlineto{\pgfpoint{-\pgfplotmarksize}{\pgfplotmarksize}}
  \pgfusepathqstroke
}
\pgfdeclareplotmark{btriangle*}{%
  \pgfpathmoveto{\pgfqpoint{0pt}{-\pgfplotmarksize}}
  \pgfpathlineto{\pgfqpointpolar{150}{\pgfplotmarksize}}
  \pgfpathlineto{\pgfqpointpolar{30}{\pgfplotmarksize}}
  \pgfpathclose
  \pgfusepathqfillstroke
}

\pgfdeclareplotmark{ltriangle}{%
  \pgfpathmoveto{\pgfqpoint{\pgfplotmarksize}{0pt}}
  \pgfpathlineto{\pgfqpointpolar{-120}{\pgfplotmarksize}}
  \pgfpathlineto{\pgfqpointpolar{120}{\pgfplotmarksize}}
  \pgfpathclose
  \pgfusepathqstroke
}
\pgfdeclareplotmark{ltriangle*}{%
  \pgfpathmoveto{\pgfqpoint{\pgfplotmarksize}{0pt}}
  \pgfpathlineto{\pgfqpointpolar{-120}{\pgfplotmarksize}}
  \pgfpathlineto{\pgfqpointpolar{120}{\pgfplotmarksize}}
  \pgfpathclose
  \pgfusepathqfillstroke
}

\pgfdeclareplotmark{rtriangle}{%
  \pgfpathmoveto{\pgfqpoint{-\pgfplotmarksize}{0pt}}
  \pgfpathlineto{\pgfqpointpolar{-60}{\pgfplotmarksize}}
  \pgfpathlineto{\pgfqpointpolar{60}{\pgfplotmarksize}}
  \pgfpathclose
  \pgfusepathqstroke
}

\definecolor{yafaxiscolor}{rgb}{0.3, 0.3, 0.3}
\definecolor{yafcolor1}{rgb}{0.4, 0.165, 0.553}
\definecolor{yafcolor2}{rgb}{0.949, 0.482, 0.216}
\definecolor{yafcolor3}{rgb}{0.47, 0.549, 0.306}
\definecolor{yafcolor4}{rgb}{0.925, 0.165, 0.224}
\definecolor{yafcolor5}{rgb}{0.141, 0.345, 0.643}
\definecolor{yafcolor6}{rgb}{0.965, 0.633, 0.267}
\definecolor{yafcolor7}{rgb}{0.627, 0.118, 0.165}
\definecolor{yafcolor8}{rgb}{0.878, 0.475, 0.686}

\pgfplotscreateplotcyclelist{yaf}{% 
{yafcolor1,mark options={scale=0.45},mark=*},
{yafcolor5,mark options={scale=0.60},mark=triangle*},
{yafcolor3,mark options={scale=0.45},mark=square*},
{yafcolor4,mark options={scale=0.70},mark=btriangle*},
{yafcolor8,mark options={scale=0.60},mark=ltriangle*},
{yafcolor7,mark options={scale=0.60},mark=rtriangle*},
{yafcolor2,mark options={scale=0.60},mark=diamond*},
{yafcolor6,mark options={scale=0.60},mark=pentagon*}}

\tikzset{
precise pin/.style args={[#1][#2]#3:#4}{
    pin={[inner sep=0pt, #1, label={[append after command={
		node [#2,
			outer sep = 0pt,
			inner sep=0pt,
			at=(\tikzlastnode),
			anchor=#3+180 ] {#4} } ]center:{}}]#3:{}}
}}

\pgfplotsset{
	clip = false,
	clip marker paths = true,
	tick align=outside,
	x tick label style = {font=\scriptsize, yshift = 1pt},
	y tick label style = {font=\scriptsize, xshift = 1pt},
	major tick length = 2pt,
    every axis y label/.style = {at = {(ticklabel cs:0.5)}, rotate=90, anchor=center, font=\scriptsize, xshift = 2pt},
	every axis x label/.style = {at = {(ticklabel cs:0.5)}, anchor=center, font=\scriptsize, yshift = -2pt},
	axis y line*=left, axis x line*=bottom,
        enlargelimits = 0.03
}
\tikzstyle{every pin}=[font=\footnotesize, inner sep = 0pt, distance=2em]
\tikzstyle{every pin edge}=[line width = 0.1pt, pin distance = 2em]

\newlength{\myheight}
\newlength{\mywidth}

\newcommand{\legDist}{Distance ($\savg{\distc{}}$)}
\newcommand{\legJacc}{Jacc.\ dist.\ ($\savg{\jacc{}}$)}
\newcommand{\legCover}{Coverage ($\cover{}$)}
\newcommand{\legDens}{Density ($\savg{\density{}}$)}

\colorlet{graphcl1}{yafcolor1!50}
\colorlet{graphcl2}{yafcolor4!30}
\colorlet{graphcl3}{yafcolor2!50}
\colorlet{graphcl4}{yafcolor5}
\colorlet{graphcl5}{yafcolor4}
\colorlet{graphcl6}{yafcolor6}

\tikzstyle{graphedge} = [black, thick, opacity = 0.5]
\tikzstyle{graphnode} = [draw = black, circle, line width = 0pt, text = black, inner sep = 0.5pt, text width = 10pt, align = center]
\tikzstyle{outliernode} = [circle, line width = 0pt, draw, text = black, fill = white, inner sep = 0.5pt, text width = 10pt, align = center]
\tikzstyle{toyedge} = [->, black, thick, bend left = 10, yafcolor5]
\tikzstyle{toynode} = [draw = black, thick, circle, line width = 0pt, text = black, inner sep = 0pt, text width = 13pt, align = center]
\tikzstyle{groupline} = [black, thick, dashed]

\makeatletter
\tikzset{multicircle/.style  args={#1, (#2)}{%
 alias=tmp@name, % 
  postaction={%
    insert path={
     \pgfextra{% 
     \pgfpointdiff{\pgfpointanchor{\pgf@node@name}{center}}%
                  {\pgfpointanchor{\pgf@node@name}{east}}%            
     \pgfmathsetmacro\insiderad{\pgf@x}%
     \foreach \c [count=\ci from = 0, evaluate=\ci as \angle using 360 - (\ci) * #1] in {#2}%
        \fill[\c] (\pgf@node@name.center)  -- ++(0:\insiderad-\pgflinewidth) arc (0:\angle:\insiderad-\pgflinewidth)--cycle;%
        }}}}}
\makeatother

\begin{document}
	\setlength{\pdfpagewidth}{8.5in}
	\setlength{\pdfpageheight}{11in}
	
	\title{\ourmaintitle}
	%\titlenote{Almost.}
	%\subtitle{through smart solution}
	%\subtitlenote{Very Smart.}
	
	\author{Alexander Marx\thanks{\rule{0pt}{1.1em}Max Planck Institute for Informatics and Saarland University, Saarbr\"{u}cken, Germany. \texttt{\{amarx,jilles\}@mpi-inf.mpg.de}} \and Jilles Vreeken\footnotemark[1]
	}
	
	\date{}
	
	\maketitle
	
\begin{abstract}
{\small\baselineskip=9pt 
{\small 

Given data over the joint distribution of two random variables $X$ and $Y$, we consider the problem of inferring the most likely causal direction between $X$ and $Y$. In particular, we consider the general case where both $X$ and $Y$ may be univariate or multivariate, and of the same or mixed data types. We take an information theoretic approach, based on Kolmogorov complexity, from which it follows that first describing the data over \textit{cause} and then that of \textit{effect} given \textit{cause} is shorter than the reverse direction. 

The ideal score is not computable, but can be approximated through the Minimum Description Length (MDL) principle. Based on MDL, we propose two scores, one for when both $X$ and $Y$ are of the same single data type, and one for when they are mixed-type. We model dependencies between $X$ and $Y$ using classification and regression trees. As inferring the optimal model is NP-hard, we propose \ourmethod, a fast greedy algorithm to determine the most likely causal direction directly from the data. 

Empirical evaluation on a wide range of data shows that \ourmethod reliably, and with high accuracy, infers the correct causal direction on both univariate and multivariate cause-effect pairs over both single and mixed-type data. }
}
\end{abstract}
	
\section{Introduction}
\label{sec:intro}

Telling cause from effect is one of the core problems in science. It is often difficult, expensive, or impossible to obtain data through randomized trials, and hence we often have to infer causality from, what is called, observational data~\cite{pearl:09:book}. We consider the setting where, given data over the joint distribution of two random variables $X$ and $Y$, we have to infer the causal direction between $X$ and $Y$. In other words, our task is to identify whether it is more likely that $X$ causes $Y$, or vice versa, that $Y$ causes $X$, or that the two are merely correlated. 

In practice, $X$ and $Y$ do not have to be of the same type. The altitude of a location (real-valued), for example, determines whether it is a good habitat (binary) for a mountain hare. In fact, neither $X$ nor $Y$ have to be univariate. Whether or not a location is a good habitat for an animal, is not just caused by a single aspect, but by a \emph{combination} of conditions, which not necessarily are of the same type. We are therefore interested in the general case where $X$ and $Y$ may be of any cardinality, and may be single or mixed-type.

To the best of our knowledge there exists no method for this general setting. Causal inference based on conditional independence tests, for example, requires three variables, and cannot decide between $\XtoY$ and $\YtoX$~\cite{pearl:09:book}. 
All existing methods that consider two variables are only defined for single-type pairs. Additive Noise Models (ANMs), for example, have only been proposed for univariate pairs of real-valued~\cite{peters:14:continuousanm} or discrete variables~\cite{peters:11:dr}, and similarly so for methods based on the independence of $P(X)$ and $P(Y\mid X)$~\cite{sgouritsa:15:cure,liu:16:dc}. 
Trace-based methods require both $X$ and $Y$ to be strictly multivariate real-valued~\cite{janzing:10:ltr,chen:13:ktr}, and whereas \ergo~\cite{vreeken:15:ergo} also works for univariate pairs, these again have to be real-valued. 
We refer the reader to Sec.~\ref{sec:rel} for a more detailed overview of related work.

Our approach is based on algorithmic information theory. That is, we follow the postulate that if $\XtoY$, it will be easier---in terms of Kolmogorov complexity---to first describe $X$, and then describe $Y$ given $X$, than vice-versa~\cite{janzing:10:algomarkov,vreeken:15:ergo,budhathoki:16:origo}. 
Kolmogorov complexity is not computable, but can be approximated through the Minimum Description Length (MDL) principle~\cite{rissanen:78:mdl,grunwald:07:book}, which we use to instantiate this framework.
In addition, we develop a causal indicator that is able to handle multivariate and mixed-type data.

To this end, we define an MDL score for coding forests, a model class where a model consists of classification and regression trees. By allowing dependencies from $X$ to $Y$, or vice versa, we can measure the difference in complexity between $\XtoY$ and $\YtoX$. Discovering a single optimal decision tree is already NP-hard~\cite{murthy:97:decision-trees}, and hence we cannot efficiently discover the coding forest that describes the data most succinctly. We therefore propose \ourmethod, an efficient greedy algorithm for discovering good models directly from data.

Through extensive empirical evaluation on synthetic, benchmark, and real-world data, we show that \ourmethod performs very well in practice. 
It performs on par with existing methods for univariate single-type pairs, is the first to handle pairs of mixed data type, and outperforms the state of the art on multivariate pairs with a large margin.
It is also very fast, taking less than 4 seconds over any pair in our experiments.

\section{Preliminaries}
\label{sec:prelim}

First, we introduce notation and give brief primers to Kolmogorov complexity and the MDL principle.

\subsection{Notation}

In this work we consider data $D$ over the joint distribution of random variables $X$ and $Y$. Such data $D$ contains $n$ records over a set $A$ of $|A| = |X| + |Y| = m$ attributes, $a_1, \dots, a_m \in A$. An attribute $a$ has a type $\textit{type}(a)$ where $\textit{type}(a) \in \{ \text{\textit{binary}, \textit{categorical}, \textit{numeric}} \}$. We will refer to binary and categorical attributes as \emph{nominal} attributes. The size of the domain of an attribute $a$ is defined as 
\begin{equation}
|\dom(a)| = \begin{cases}
\#\textit{values} &\text{if \textit{type}$(a)$ is nominal}\\
\frac{\max(a) - \min(a)}{\res(a)} + 1&\text{if \textit{type}$(a)$ is numeric} \; ,
\end{cases}
\end{equation}
where $\res(a)$ is the resolution at which the data over attribute $a$ was recorded. For example, a resolution of 1 means that we consider integers, of $0.01$ means that $a$ was recorded with a precision of up to a hundredth.

We will consider decision and regression trees. In general, a tree $T$ consist of $|T|$ nodes. We identify internal nodes as $\node \in \internal(T)$, and leaf nodes as $\leaf \in \lvs(T)$. A leaf node $l$ contains $|l|$ data points. 

All logarithms are to base 2, and we use $0 \log 0 = 0$.

\subsection{Kolmogorov Complexity, a brief primer}

The Kolmogorov complexity of a finite binary string $x$ is the length of the shortest binary program $p^*$ for a universal Turing machine $\mathcal{U}$ that generates $x$, and then halts~\cite{kolmogorov:65:information, vitanyi:93:book}. Formally, we have
\[
K(x) = \min \{ |p| \mid p \in \{0,1\}^*, \mathcal{U}(p) = x \} \; .
\]
Simply put, $p^*$ is the most succinct \emph{algorithmic} description of $x$, and the Kolmogorov complexity of $x$ is the length of its ultimate lossless compression. Conditional Kolmogorov complexity, $K(x \mid y) \leq K(x)$, is then the length of the shortest binary program $p^*$ that generates $x$, and halts, given $y$ as input. For more details see~\cite{vitanyi:93:book}.

\subsection{MDL, a brief primer}

The Minimum Description Length (MDL) principle~\cite{rissanen:78:mdl, grunwald:07:book} is a practical variant of Kolmogorov Complexity. Intuitively, instead of all programs, it considers only those programs that we know that output $x$ and halt. Formally, given a model class $\models$, MDL identifies the best model $M \in \models$ for data $\data$ as the one minimizing 
\[
L(\data, M) = L(M) + L(\data \mid M) \; ,
\]
where $L(M)$ is the length in bits of the description of $M$, and $L(\data\mid\model)$ is the length in bits of the description of data $\data$ given $M$. This is known as two-part MDL. There also exists one-part, or \emph{refined} MDL, where we encode data and model together. Refined MDL is superior in that it avoids arbitrary choices in the description language $L$, but in practice only computable for certain model classes. Note that in either case we are only concerned with code \emph{lengths} --- our goal is to measure the \emph{complexity} of a dataset under a model class, not to actually compress it~\cite{grunwald:07:book}. 
	
\section{Causal Inference by Compression}
\label{sec:causal}

We pursue the goal of causal inference by compression. Below we give a short introduction to the key concepts. 

\subsection{Causal Inference by Complexity}

The problem we consider is to infer, given data over two correlated variables $X$ and $Y$, whether $X$ caused $Y$, whether $Y$ caused $X$, or whether $X$ and $Y$ are only correlated. As is common, we assume causal sufficiency. That is, we assume there exists no hidden confounding variable $Z$ that is the common cause of both $X$ and $Y$.

The Algorithmic Markov condition, as recently postulated by Janzing and Sch\"{o}lkopf~\cite{janzing:10:algomarkov}, states that factorizing the joint distribution over \cause and \effect into $P(\cause)$ and $P(\effect \mid \cause)$, will lead to simpler---in terms of Kolmogorov complexity---models than factorizing it into $P(\effect)$ and $P(\cause \mid \effect)$. Formally, they postulate that if $X$ causes $Y$, 
\begin{equation}
K(P(X)) + K(P(Y \mid X)) \le K(P(Y)) + K(P(X \mid Y)) \; . \label{eq:janzing}
\end{equation}
While in general the symmetry of information, $K(x)+K(y\mid x) = K(y) + K(x \mid y)$, holds up to an additive constant~\cite{vitanyi:93:book}, Janzing and Sch\"{o}lkopf~\cite{janzing:10:algomarkov} showed it does \emph{not} hold when $X$ causes $Y$, or vice versa. Based on this, Budhathoki \& Vreeken~\cite{budhathoki:16:origo} proposed 
\begin{equation}
\DXY^{*} = \frac{K(P(\dataX)) + K(P(\dataY \mid \dataX))}{K(P(\dataX)) + K(P(\dataY))} \; , \label{eq:origo}
\end{equation}
as a causal indicator that uses this asymmetry to infer that $X \rightarrow Y$ as the most likely causal direction if $\DXY^* < \DYX^*$, and vice versa. The normalisation has no function during inference, but does help to interpret the confidence of the indicator.

Both scores assume access to the true distribution $P(\cdot)$, whereas in practice we only have access to empirical data. Moreover, following from the halting problem, Kolmogorov complexity is not computable. We can approximate it, however, via MDL~\cite{vitanyi:93:book,grunwald:07:book}, which also allows us to directly work with empirical distributions.

\subsection{Causal Inference by MDL}

For causal inference by MDL, we will need to approximate both $K(P(\dataX))$ and $K(P(\dataY \mid \dataX))$. For the former, we need to consider the model classes $\models_{X}$ and $\models_{Y}$, while for the latter we need to consider class $\models_{Y\mid X}$ of models $\model_{Y \mid X}$ that describe the data of $Y$ dependent the data of $X$. 

That is, we are after the \emph{causal} model $\model_{\XtoY}= (\model_{X}, \model_{Y\mid X})$ 
from the class $\models_\XtoY = \models_{X} \times \models_{Y \mid X}$ that best describes the data $Y$ by exploiting as much as possible structure of $X$ to save bits. By MDL, we identify the optimal model $\model_\XtoY \in \models_\XtoY$ for data $\data$ over $X$ and $Y$ as the one minimizing
\[
L(\data,\model_\XtoY) = L(\dataX, \model_{X}) + L(\dataY, M_{Y \mid X} \mid \dataX) \; ,
\]
where the encoded length of the data of $X$ under a given model is encoded using two-part MDL, similarly so for $Y$, if we consider the inverse direction.

To identify the most likely causal direction between $X$ and $Y$ by MDL we can now simply rewrite Eq.~\eqref{eq:origo}
\[
\DXY = \frac{L(\dataX, \model_{X}) + L(\dataY, M_{Y\mid X} \mid X)}{L(\dataX, \model_{X}) + L(\dataY, \model_{Y})} \; .
\]
Similar to the original score, we infer that $X$ is a likely cause of $Y$ if $\DXY < \DYX$, $Y$ is a likely cause of $X$ if $\DYX < \DXY$, and that $X$ and $Y$ are only correlated or might have a common cause if $\DXY = \DYX$.

\subsection{Normalized Causal Indicator}

Although $\Delta$ has nice theoretical properties, it has a mayor drawback. It assumes that a bit gain in the description of the data over one attribute has the same importance as one bit gain in the description of the data over another attribute. This does not hold true if these attributes have different intrinsic complexities, such as when their domain sizes strongly differ. For example, a continuous valued attribute is very likely to have a much higher intrinsic complexity than a binary attribute. This means that gaining $k$ bits from an attribute with a large domain is not comparable to gaining $k$ bits from an attribute with a small domain. Since the $\Delta$ indicator compares the absolute difference in bits, it does not account for differences w.r.t. the intrinsic complexity. Hence, $\Delta$ is highly likely to be a bad choice when $X$ and $Y$ are of different, or of mixed-type data.

We therefore propose an alternative indicator for causal inference on mixed-type data. Instead of taking the absolute difference between the conditioned and unconditioned score, we instead consider relative differences w.r.t. the marginal. We can derive the \textit{Normalized Causal Indicator} (\NCI) starting from the numerator of the $\Delta$ indicator. By subtracting the conditional costs on both sides, we have
\[
L(\dataX, \model_{X}) - L(\dataX, M_{X| Y} | Y) < L(\dataY, \model_{Y}) - L(\dataY, M_{Y| X} | X).
\]
Since the aim of the \NCI is to measure the relative gain, we divide by the costs of the unconditioned data
\[
\frac{L(\dataX, \model_{X}) - L(\dataX, M_{X| Y} | Y)}{L(\dataX, \model_{X})} = 1 - \frac{L(\dataX, M_{X| Y} | Y)}{L(\dataX, \model_{X})} \; .
\]
After this step, we can conclude that for the relative gain it holds, if $\XtoY$
\[
\frac{L(\dataX, M_{X\mid Y} \mid Y)}{L(\dataX, \model_{X})} > \frac{L(\dataY, M_{Y\mid X} \mid X)}{L(\dataY, \model_{Y})} \; .
\]
This score can be understood as an instantiation of the \ergo indicator proposed by Vreeken~\cite{vreeken:15:ergo}. From the derivation, we can easily see that the difference between the score of both indicators depends only on the normalization factor and hence both are based on the Algorithmic Markov condition. It turns out, however, that the \ergo indicator is also biased. Although it balances the gain between $X$ and $Y$, we need a score that does not impose prior assumptions to the individual attributes of $X$ and $Y$. With the \ergo indicator, it could happen that a single $X_i \in X$ dominates the whole score for $X$. To account for this, we assume independence among the variables within $X$ and $Y$, meaning that the domain of two individual attributes within $X$ or $Y$ is allowed to differ. Hence, we formulate the \NCI, which we from now on denote by $\delta$, from $X$ to $Y$ as
\[
\delta_{\XtoY} = \frac{1}{|Y|} \sum_{Y_i \in Y} \frac{L(Y_i, M_{Y_i \mid X} \mid X)}{L(Y_i, \model_{Y_i})}\;
\]
and analogously $\delta_{\YtoX}$. To avoid bias towards dimensionality, we normalize by the number of attributes. As above, we infer $\XtoY$ if $\delta_{\XtoY} < \delta_{\YtoX}$ and vice versa.

In practice, we expect that $\Delta$ performs well on data where $X$ and $Y$ are of the same type, especially when $|X|=|Y|$ and the domain sizes of their attributes are balanced. For unbalanced domains, dimensionality, and especially for mixed-type data, we expect $\delta$ to perform much better. The experiments indeed confirm this.
	
\section{MDL for Tree Models}
\label{sec:score}

To use the above defined causal indicators in practice, we need to define a casual model class $\models_\XtoY$, how to encode a model $\model \in \models$ in bits, and how to encode a dataset $\data$ using a model $\model$. As models we consider tree models, or, \emph{coding forests}. 
A coding forest $\model$ contains per attribute $a_i \in A$ one coding tree $T_i$. A coding tree $T_i$ encodes the values of $a_i$ in its leaves, splitting or regressing the data of $a_i$ on attribute $a_j$ ($i \neq j$) in its internal nodes to encode the data of $a_i$ more succinctly. 

We encode the data over attribute $a_i$ with the corresponding coding tree $T_i$. The encoded length of data $D$ and $\model$ then is $L(D, \model) = \sum_{a_i \in \attributes} L(\tree_i)$, which corresponds to the sum of costs of the individual trees.

To ensure lossless decoding, there needs to exist an order on the trees $T \in \model$ such that we can transmit these one by one. In other words, in a \emph{valid} tree model there are no cyclic dependencies between the trees $\tree \in \model$, and a valid model can hence be represented by a DAG. Let $\models(D)$ be the set of all valid tree models for $D$, that is, $\model \in \models(D)$ is a set of $|A|$ trees such that the data types of the leafs in $\tree_i$ corresponds to the data type of attribute $a_i$, and its dependency graph is acyclic.

\begin{figure}[t]
\begin{minipage}[t]{.5\linewidth}
\centering
\includegraphics[]{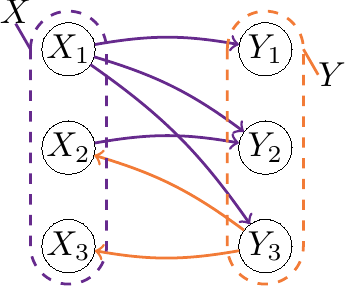}
\subcaption{DAG}\label{toy:dag}
\end{minipage}%
\begin{minipage}[t]{.5\linewidth}
\centering
\includegraphics[]{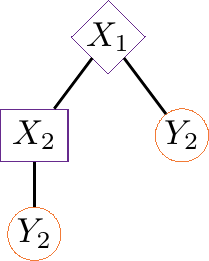}
\subcaption{Tree for $Y_2$}\label{toy:tree}
\end{minipage}%
\caption{Toy data set with ground truth $\XtoY$. Shown is the dependency DAG (right). More dependencies go from $X$ to $Y$ than vice versa. Left: Example coding tree for $Y_2$. $X_1$ splits the values of $Y_2$ into two subsets. In addition, the subset belonging to the left child can be further compressed by regressing on $X_2$.}
\label{fig:toy:example}
\end{figure}

We write $\models_{X}(X)$ and $\models_{Y}(Y)$ to denote the subset of valid coding forests for $X$ and $Y$, where we do not allow dependencies. To describe the possible set of models where we allow attributes of $X$ to only depend on attributes of $Y$ we write $\models_{X \mid Y}(X)$ and do so accordingly for $Y$ depending only on $X$. If an attribute does not have any incoming dependencies, its tree is a stump. Fig.~\ref{fig:toy:example} shows the DAG for a toy data set, and an example tree for $Y_2$. From the DAG, the set of purple edges would be a valid model in $\models_{Y \mid X}(Y)$, whereas the orange edges are a valid model from $\models_{X \mid Y}(X)$. 

\subsubsection*{Cost of a Tree}
The encoded cost of a tree consists of two parts. First, we transmit the topology of the tree. From the root node on we indicate with one bit per node whether it is a leaf or an internal node, and if the latter, one further bit to identify whether it is a split or regression node. Formally we have that
\[
L(\tree) = |\tree| + \sum_{\node \in \internal(\tree)} (1+L(\node)) + \sum_{\leaf \in \lvs(\tree)} L(\leaf) \; .
\]
Next, we explain how we encode internal nodes and then specify the encoding for leaf nodes.

\subsubsection*{Cost of a Single Split}
The length of a split node $\node$ is
\[
L_{1\tsplit}(\node) = 1 + \log |\attributes| + \begin{cases}
\log |\dom(a_j)| \text{ if $a_i$ is categorical,}\\
\log |\dom(a_j) - 1| \text{ else.}
\end{cases}
\]
whereas we first 
need one bit to indicate this is a single-split node, then
identify in $\log |\attributes|$ bits on which attribute $a_j$ we split, 
and third the split condition.

The split condition can be any value in the domain for categorical, and can lie in between two consecutive values of a numeric attribute ($|\dom(a_j) - 1|$ choices). For binary we only have one option, resulting in zero cost.

\subsubsection*{Costs of a Multiway split}
A multiway split is only possible for categorical and real valued data. As there are exponentially many multiway splits, we consider only a subset. The costs for a multiway split are
\[
L_{\text{k}\tsplit}(\node) = 1 + \log |\attributes| + \begin{cases}
0 \text{ if $a_i$ is categorical,}\\
L_{\mathbb{N}}(k) \text{ numeric,}
\end{cases}
\]
where the first two terms are similar to above. For categorical data, we only consider splitting on all values, and hence have no further cost. For numeric data, we only split non-deterministic cases, i.e. if there exist duplicate values. To do so, we split on every such value that occurs at least $k$ times, and one residual split for all remaining data points.
To encode such a split, we transmit $k$ using $L_{\mathbb{N}}(k)$ bits, where $L_{\mathbb{N}}$ is the MDL optimal encoding for integers $z \geq 1$~\cite{rissanen:83:integers}. 

\subsubsection*{Cost of Regressing}

For a regression node we also first encode the target attribute, and then the parameters of the regression, i.e.
\[
L_{\tregress}(\node) = \log |\attributes| + \sum_{\phi \in \Phi(\node)} \left( \, 1 + \LN(s) + \LN(\lfloor \phi \cdot 10^{s}\rfloor) \,  \right) ,
\]
where $\Phi(\node)$ denotes the set of parameters for the regression. For linear regression, it consists of $\alpha$ and $\beta$, while for quadratic regression it further contains $\gamma$. To describe each parameter $\phi \in \Phi$ we transmit it up to a user defined precision, e.g. $0.001$, we first encode the corresponding number of significant digits $s$, e.g. $3$, and then the shifted parameter value.

Next, we describe how to encode the data in a leaf $l$. As we consider both nominal and numeric attributes, we need to define $L_\nom(l)$ for nominal and $L_\num(l)$ for numeric data.

\subsubsection*{Cost of a Nominal Leaf}
To encode the data in a leaf of a nominal attribute, we can use refined MDL~\cite{kontkanen:07:histo}. That is, we encode minimax optimally, without having to make design choices~\cite{grunwald:07:book}. In particular, we encode the data of a nominal leaf using the normalized maximum likelihood (NML) distribution as
\begin{align}
L_\nom(\leaf) =& \log \left( \sum_{\substack{h_1 + \cdots + h_{k} = |\leaf|}} \frac{|\leaf|!}{h_1! h_2! \cdots h_{k}!} \right)\\
& - |l| \sum_{c \in \dom(a_i)} \Pr(a_i = c \mid \leaf) \log \Pr(a_i = c \mid \leaf) \; .
\end{align}
Kontkanen \& Myllym\"{a}ki~\cite{kontkanen:07:histo} derived a recursive formula to calculate this in linear time. 

\subsubsection*{Cost of a Numerical Leaf}

For numeric data existing refined MDL encodings have high computational complexity \cite{kontkanen:07:histo}. Hence, we encode the data in numeric leaves using two-part MDL, using point models with Gaussian or uniform noise. A split or a regression on an attribute aims to reduce the variance or the domain in the leaf. We encode the costs of a numeric leaf as 
\begin{align}
L_\num(\leaf \mid \sigma, \mu) =& \frac{|l|}{2} \left( \frac{1}{\ln 2} + \log 2 \pi \sigma^2 \right) - |l| \log \res(a_i) ,
\end{align}
given empirical mean $\mu$ and variance $\sigma$ or as uniform given $\min$ and $\max$ as
\begin{align}
L_\num(\leaf \mid \min, \max) =& |l| \cdot \log \left( \frac{\max - \min}{\res(a_i)} + 1 \right) \; .
\end{align}
We encode the data as Gaussian if this costs fewer bits than encoding it as uniform. To indicate this decision, we use one bit and encode the minimum of both plus the corresponding parameters.
As we consider empirical data, we can safely assume that all parameters lie in the domain of the given attribute. Since we do not have any preference on the parameter values, the encoded costs of a numeric leaf $\leaf$  are 
\begin{align}
L_\num(l) &= 1 + 2 \log |\dom(a_j)| \\
&+ \min \{ L_\num(l \mid \sigma, \mu), L_\num(\leaf \mid \min, \max) \} \; .
\end{align}

Putting it all together, we now know how to compute $L(D,\model)$, by which we can formally define the Minimal Coding Forest problem.

\vspace{0.5em}
\noindent\textbf{Minimal Coding Forest Problem} 
\emph{
Given a data set $\data$ over a set of attributes $A = \{a_1, \ldots, a_m\}$, and $\models$ a valid model class for $A$. Find the smallest model $\model \in \models$ such that $L(\data,\model)$ is minimal.
}
\vspace{0.5em}

From the fact that both inferring optimal decision trees and structure learning of Bayesian networks---to which our tree-models reduce for nominal-only data and splitting on all values---are NP-hard~\cite{murthy:97:decision-trees}, it trivially follows that the Minimal Coding Forest problem is also NP-hard. Hence, we resort to heuristics.

\section{The \ourmethod Algorithm}
\label{sec:algo}

Knowing the score $L(\data,\model)$ and the problem, we can now introduce the \ourmethod algorithm, which stands for \textbf{c}lassification and \textbf{r}egression based p\textbf{ack}ing of data. \ourmethod is an efficient greedy heuristic for discovering a coding forest $\model$ from given model class $\models$ with low $L(\data,\model)$. It builds upon the well-known ID3 algorithm~\cite{quinlan:86:id3}. In the next section we explain the main aspects of the algorithm.

\subsubsection*{Greedy algorithm}
We give the pseudocode of \ourmethod as Algorithm \ref{alg:crack}. Before running the algorithm, we set the resolution per attribute, which is $1$ for nominal data (line~\ref{alg:crack:res}). For numeric data, we calculate the differences between adjacent values, and to reduce sensitivity to outliers take the $k^{\mathit{th}}$ smallest difference as resolution. In general, setting $k$ to $0.1n$ works well in practice.

\GreedyCrack starts with an empty model consisting of only trivial trees, i.e. leaf nodes containing all records, per attribute (line~\ref{alg:crack:trivialtrees}). The given model class $\models$ implicitly defines a graph $\depgraph$ of dependencies between attributes that we are allowed to consider (line~\ref{alg:crack:depgraph}). 
To make sure the returned model is valid, we need to maintain a graph representing its dependencies (lines~\ref{alg:crack:checkgraph1}--\ref{alg:crack:checkgraph2}).
We iteratively discover that refinement of the current model that maximizes compression. To find the best refinement, we consider every attribute (line~\ref{alg:crack:everyattrib}), and every legal additional split or regression of its corresponding tree (line~\ref{alg:crack:splitregress}). A refinement is only legal when the dependency is allowed by the model family (line~\ref{alg:crack:depcheck}), the dependency graph remains acyclic, and we do not split or regress twice on the same attribute (line~\ref{alg:crack:validcheck}). We keep track of the best found refinement.

The key subroutine of \ourmethod is \SplitOrRegress, in which we discover the optimal refinement of a leaf $l$ in tree $T_i$. That is, it finds the optimal split of $l$ over all candidate attributes $a_j$ such that we minimize the encoded length. In case both $a_i$ and $a_j$ are numeric, \SplitOrRegress also considers the best linear and quadratic regression and decides for the variant with the best compression---choosing to split in case of a tie. In the interest of efficiency, we do not allow splitting or regressing multiple times on the same candidate. 

\begin{algorithm}[tb!]
	\caption{$\GreedyCrack(\data, \models)$}
	\label{alg:crack}
	\Input{ data $\data$ over attributes $A$, model class $\models$}
	\Output{ tree model $\model \in \models$ with low $L(\data, \model)$}
	
	$\res(a_i) \leftarrow \RobustMinDiff(a_i)$\; \label{alg:crack:res} 
	$\tree_i \leftarrow \TrivialTree(a_i)$ for all $a_i \in A$\; \label{alg:crack:trivialtrees}
	$\depgraph \leftarrow$ dependency graph for $\models$\; \label{alg:crack:depgraph}
	$V \leftarrow \{ v_i \mid i \in A \}, \; E \leftarrow \emptyset$\; \label{alg:crack:checkgraph1}
	$\depgraph \leftarrow (V,E)$\; \label{alg:crack:checkgraph2}
	\While{$L(\data,\model)$ decreases}{
		\For {$a_i \in A$}{\label{alg:crack:everyattrib}
			$O_i \leftarrow \tree_i$\;
			\For{$l \in \lvs(\tree_i), (i,j) \in \depgraph$}{ \label{alg:crack:depcheck}
				\If{$E \cup (v_i, v_j) \text{ is acyclic}$ \AND $j \notin \text{path}(l)$}{ \label{alg:crack:validcheck}
					$\tree'_i \leftarrow \SplitOrRegress(\tree_i, l, j)$\;\label{alg:crack:splitregress}
					\If{$L(\tree'_i) < L(O_i)$}{\label{alg:crack:bettercheck}
						$O_i \leftarrow \tree'_i, \; e_i \leftarrow j$\;\label{alg:crack:betterstore}
					}
				}	
			}
		}
		$k \leftarrow \arg\min_i \{ L(O_i) - L(\tree_i) \}$\;\label{alg:crack:bestk}
		\If{$L(O_k) < L(\tree_k)$}{\label{alg:crack:bestcheck}
			$\tree_k \leftarrow O_k$\;\label{alg:crack:beststore}
			$E \leftarrow E \cup (v_k, v_{e_k})$\label{alg:crack:bestedge}
		}
	}
	\Return{$\model \leftarrow \bigcup_i \tree_i $} \label{alg:crack:return}
\end{algorithm}	

Since we use a greedy heuristic to construct the coding trees, we have a worst case runtime of $O(2^{m}n)$, where $m$ is the number of attributes and $n$ is the number of rows. Although the worst case runtime is exponential, in practice, \ourmethod takes only a few seconds.

\subsubsection*{Causal Inference with \GreedyCrack}

To compute our causal indicators we have to run \ourmethod twice on $D$. First with model class $\models_{X \mid Y}(X)$ to obtain $\model_{X \mid Y}(X)$ and second with $\models_{Y\mid X}(Y)$, to obtain $\model_{Y\mid X}(Y)$. To estimate $\models_{X}(X)$, we assume a uniform prior $L(X \mid \model_{X}) = -n\sum_{a_i \in X} \log res(a_i)$ and similarly for $\model_{Y}(Y)$. We can use these scores to calculate both the $\delta$ score and the $\Delta$ score. We will refer to \ourmethod using the $\delta$ indicator as \ourmethode, and \ourmethod with the $\Delta$ indicator as \ourmethodo.
	
\section{Related Work}
\label{sec:rel}

Causal inference on observational data is a challenging problem, and has recently attracted a lot of attention~\cite{pearl:09:book, janzing:10:algomarkov, shimizu:06:anm, budhathoki:16:origo}. Most existing proposals, however, are highly specific in the type of causal dependencies and type of variables they can consider. 

Clasical constrained-based approaches, such as conditional independence tests, require three observed random variables~\cite{spirtes:00:book,pearl:09:book}, cannot distinguish Markov equivalent causal DAGs~\cite{verma:90:markov-equiv} and hence cannot decide between $\XtoY$ and $\YtoX$. Recent approaches use properties of the joint distribution to break the symmetry.

Additive Noise Models (ANMs)~\cite{shimizu:06:anm}, for example, assume that the effect is a function of the cause and cause-independent additive noise. ANMs exist for univariate real-valued~\cite{shimizu:06:anm,hoyer:09:nonlinear,zhang:09:ipcm,peters:14:continuousanm} and discrete data~\cite{peters:10:discreteanm}. 
A related approach considers the asymmetry in the joint distribution of $\cause$ and $\effect$ for causal inference. The linear trace method (\ltr)~\cite{janzing:10:ltr} and the kernelized trace method (\ktr)~\cite{chen:13:ktr} aim to find a structure matrix $A$ and the covariance matrix $\Sigma_X$ to express $Y$ as  $AX$. Both methods are restricted to multivariate continuous data. 
Sgouritsa et al.~\cite{sgouritsa:15:cure} show that the marginal distribution $P(\cause)$ of the cause is independent of the conditional distribution $P(\effect \mid \cause)$ of the effect. 
They proposed \cure, using unsupervised reverse regression on univariate continuous pairs. Liu et al~\cite{liu:16:dc} use distance correlation to identify the weakest dependency between univariate pairs of discrete data. 

The algorithmic information-theoretic approach views causality in terms of Kolmogorov complexity. The key idea is that if $X$ causes $Y$, the shortest description of the joint distribution $P(X, Y)$ is given by the separate descriptions of the distributions $P(X)$ and $P(Y \mid X)$~\cite{janzing:10:algomarkov}, and justifies additive noise model based causal inference~\cite{janzing:10:justifyanm}.
However, as Kolmogorov complexity is not computable~\cite{vitanyi:93:book}, causal inference using algorithmic information theory requires practical implementations, or notions of independence. For instance, the information-geometric approach~\cite{janzing:12:igci} defines independence via orthogonality in information space for univariate continuous pairs. 
Vreeken~\cite{vreeken:15:ergo} instantiates it with the cumulative entropy to infer the causal direction in continuous univariate and multivariate data. Mooij instantiates the first practical compression-based approach~\cite{mooij:10:mml} using the Minimum Message Length. Budhathoki and Vreeken approximate $K(X)$ and $K(Y \mid X)$ through MDL, and propose \origo, a decision tree based approach for causal inference on multivariate binary data~\cite{budhathoki:16:origo}. Marx and Vreeken\cite{marx:17:slope} proposed \slope, an MDL based method employing local and global regression for univariate numeric data.

In contrast to all methods above, \ourmethod can consider pairs of any cardinality, univariate or multivariate, and of same, different, or even mixed-type data.

\section{Experiments}
\label{sec:exps}

In this section, we evaluate \ourmethod empirically. We implemented \ourmethod in C++, and provide the source code including the synthetic data generator along with the tested datasets for research purposes.\!\footnote{\codeurl} The experiments concerning \ourmethod were executed single-threaded.
All tested data sets could be processed within seconds; over all pairs the longest runtime for \ourmethod was $3.8$ seconds.

We compare \ourmethod to \cure~\cite{sgouritsa:15:cure}, \igci~\cite{janzing:12:igci}, \ltr~\cite{janzing:10:ltr}, \origo~\cite{budhathoki:16:origo}, \ergo~\cite{vreeken:15:ergo} and \slope~\cite{marx:17:slope} using their publicly available implementations and recommended parameter settings.

\subsection{Synthetic data}

The aim of our experiments on synthetic data is to show the advantages of either score. In particular, we expect \ourmethodo to perform well on nominal data and numeric data with balanced domain sizes and dimensions. On the other hand, \ourmethode should have an advantage when it comes to numeric data with varying domain sizes and mixed-type data.

We generate synthetic data with assumed ground truth $\XtoY$ with $|X| = k$ and $|Y| = l$, each having $n=5 \, 000$ rows, in the following way. First, we randomly assign the type for each attribute in $X$. For nominal data, we randomly draw the number of classes between two (binary) and five and distribute the classes uniformly. Numeric data is generated following a normal distribution taken to the power of $q$ by keeping the sign, leading to a sub-Gaussian ($q < 1.0$) or super-Gaussian ($q > 1.0$) distribution.\!\footnote{We use super- and sub-Gaussians to ensure identifiability.}

To create data with the true causal direction $\XtoY$, we introduce dependencies from $X$ to $Y$, where we distinguish between splits and refinements. We call the probability threshold to create a dependency $\varphi \in [ 0, 1 ]$. For each $j \in \{ 1, \dots, l \}$, we throw a biased coin based on $\varphi$ for each $X_i \in X$ that determines if we model a dependency from $X_i$ to $Y_j$. A split means that we find a category (nominal) or a split-point (numeric) on $X_i$ to split $Y_j$ into two groups, for which we model its distribution independently. As refinement, we either do a multiway split or model $Y_j$ as a linear or quadratic function of $X_i$ plus independent Gaussian noise. 

\paragraph{Accuracy}

\begin{figure}[t]%
	\begin{minipage}[t]{.42\linewidth}
	\includegraphics[]{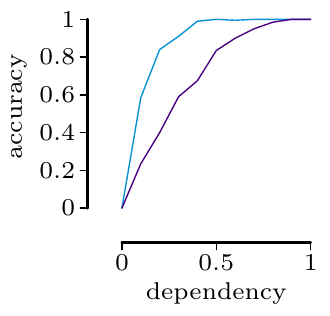}
	\subcaption{nominal}\label{fig:dep:nom}
	\end{minipage}%
	\begin{minipage}[t]{.29\linewidth}
	\includegraphics[]{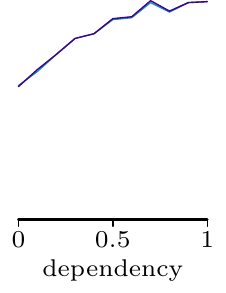}
	\subcaption{numeric}\label{fig:dep:num}
	\end{minipage}%
	\begin{minipage}[t]{.29\linewidth}
	\includegraphics[]{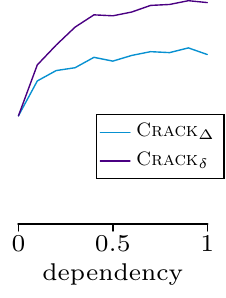}
	\subcaption{mixed}\label{fig:dep:mixed}
	\end{minipage}%
	\caption{Accuracy for $\Delta$ and $\delta$ on nominal, numeric and mixed-type data based on the dependency.}
	\label{fig:dependency}
\end{figure}

First, we compare the accuracies of \ourmethode and \ourmethodo with regard to single-type and mixed-type data. To do so, we generate $200$ synthetic data sets with $|X| = |Y| = 3$ for each dependency level where $\varphi \in \{0.0,0.1,\dots 1.0 \}$. Figure~\ref{fig:dependency} shows the results for numeric, nominal and mixed-type data.
At $\varphi = 1.0$ both approaches reach nearly $100\%$ accuracy on single-type data. For single-type data, the accuracy of both methods increases with the dependency. At $\varphi = 0$, both approaches correctly do not decide instead of taking wrong decisions. 
As expected \ourmethode strongly outperforms \ourmethodo on mixed-type data, reaching near $100\%$ accuracy, whereas \ourmethodo reaches only $72\%$. On nominal data, \ourmethodo picks up the correct signal faster than \ourmethode.

\paragraph{Dimensionality}

Next, we check how sensitive both scores are to dimensionality, whereas we discriminate between asymmetric $k \neq l$ and symmetric $k=l$. We evaluated $200$ data sets per dimensionality. For the symmetric case, both methods are near to $100\%$ on single-type data, whereas only \ourmethode also reaches this target on mixed-type data, as can be seen in the appendix.\!\footnotemark[1] We now discuss the more interesting case for asymmetric pairs in detail.

\begin{figure}[t]
	\hfill
	\includegraphics[]{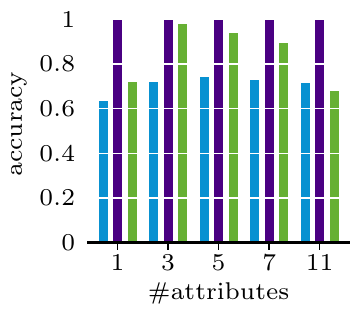}
	\hfill
	\includegraphics[]{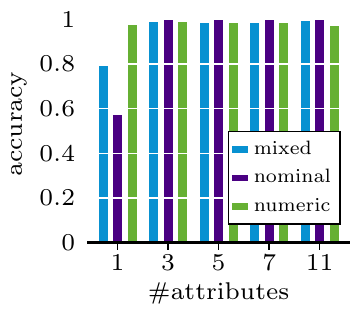}
	\hfill
	\caption{Accuracy of $\Delta$ (left) and $\delta$ (right) on asymmetric dimensions $k\in \{1,3,5,7,11\}$ and $3$ for nominal, numeric and mixed-type data.}
	\label{fig:asymmetric}
\end{figure}

To test asymmetric pairs, we keep the dimension of one variable at three, $k = 3$, while we increase the dimension of the second variable $l$ from one to eleven. To avoid bias, we assigned the dimension $k$ to $X$ and $l$ to $Y$ and swap the dimensions in every other test. We show the results in Figure~\ref{fig:asymmetric}. As expected, we observe that \ourmethode has much fewer difficulties with the asymmetric data sets than \ourmethodo. From $l = 3$ onwards, \ourmethode is close to $100\%$. On nominal data, \ourmethodo performs near perfect and also has the clear advantage for $l=1$.

\subsection{Real world data}

Based on the evaluation on synthetic data, we test our approach on univariate benchmark data and multivariate data consisting of known test sets and new causal pairs with known ground truth that we present in the current paper.

\paragraph{Univariate benchmark}

\begin{figure}[t]
	\centering
	\includegraphics[]{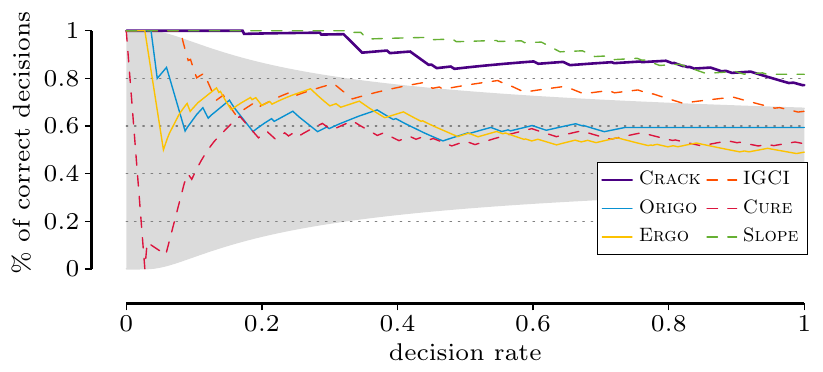}
	\caption{[Higher is better] Decision rates of \ourmethod, \origo, \igci, \cure, \ergo and \slope on univariate Tuebingen pairs (100) weighted as defined. Approaches that are only applicable to univariate data are drawn with dotted lines.}
	\label{fig:decision_rate}
\end{figure}

To evaluate \ourmethod on univariate data, we apply it to the well-known Tuebingen benchmark data set that consists of $100$ univariate pairs.\!\footnote{https://webdav.tuebingen.mpg.de/cause-effect/} The pairs mainly consist of numeric data and a few categoric instances. Therefore, we apply \ourmethodo. We compare to the state of the art methods that are applicable to multivariate and univariate data, \origo~\cite{budhathoki:16:origo} and \ergo~\cite{vreeken:15:ergo}, and methods specialized for univariate pairs, \cure~\cite{sgouritsa:15:cure}, \igci~\cite{janzing:12:igci} and \slope~\cite{marx:17:slope}. For each approach, we sort the results by their confidence. According to this order, we calculate for each position $k$ the percentage of correct inferences up to this point, called the decision rate. We weigh the decisions as specified by the benchmark, plot the results in Fig.~\ref{fig:decision_rate} and show the $95\%$ confidence interval of a fair coin flip as a grey area. Except to \ourmethod and \slope, all methods are insignificant w.r.t. the fair coin flip. In particular, \ourmethod has an accuracy of over $90\%$ for the first $41\%$ of its decisions and reaches $77.2\%$ overall.
Regarding the whole decision rate, \ourmethod is nearly on par with \slope, which is as far as we know, the current state of the art for univariate continuous data.

\paragraph{Multivariate data}

To test \ourmethode on multivariate mixed-type and single-type data, we collected 17 data sets. The information of the dimensionality for each data set is listed in Table~\ref{tab:mv_comparison}. The first six data sets belong to the Tuebingen benchmark data set~\cite{mooij:16:pairs} and the next four were published by Janzing et al.~\cite{janzing:10:ltr}. Further, we extracted cause-effect pairs form the \textit{Haberman}~\cite{haberman:76:dataset}, \textit{Iris}~\cite{fisher:36:dataset}, \textit{Mammals}~\cite{heikinheimo:07:mammals} and \textit{Octet}~\cite{ghiringhelli:15:octet, vechten:69:quantum} data sets. Those are described in more detail in the appendix.

We compare \ourmethode with \ltr, \ergo and \origo. \ergo and \ltr do not consider categoric data, and are hence
not applicable on all data sets. In addition, \ltr is only applicable to strictly multivariate data sets. \ourmethode is applicable to all data sets, infers $15/17$ causal directions correctly, by which it has an overall accuracy of $88.2\%$. Importantly, the two wrong decisions have low confidences compared to the correct inferences.

\begin{table}[t]
\centering
\newcommand{\na}{\tiny{(n/a)}}
	\begin{tabular}{l@{\hspace{0.6em}} r@{\hspace{0.5em}} r@{\hspace{0.4em}} r c@{\hspace{0.35em}} c@{\hspace{0.35em}} c@{\hspace{0.35em}} c}
		\toprule
		& & & & & \multicolumn{3}{l}{\textbf{Decisions}} \\ 
		\cmidrule{5-8}
		\textbf{Dataset}&$m$&$k$&$l$&\footnotesize\ltr&\footnotesize\ergo&\footnotesize\origo&\footnotesize\ourmethod \\
		\midrule
		Climate&$10\,226$&4&4&\cmark&\cmark&--&-- \\
		Ozone&$989$&1&3&\na&\cmark&\cmark&\cmark \\
		Car&$392$&3&2&--&\cmark&\cmark&\cmark \\
		Radiation&$72$&16&16&--&--&--&\cmark \\
		Symptoms&$120$&6&2&\cmark&\cmark&--&\cmark \\
		Brightness&$1\,000$&9&1&\na&\na&--&\cmark \\
		Chemnitz&$1\,440$&3&7&\cmark&\cmark&\cmark&\cmark \\
		Precip.&$4\,748$&3&12&\cmark&--&--&\cmark \\
		Stock 7&$2\,394$&4&3&--&\cmark&--&\cmark \\
		Stock 9&$2\,394$&4&5&--&\cmark&--&\cmark \\
		Haberman&$306$&3&1&\cmark&\cmark&--&--\\
		Iris flower&$150$&4&1&\na&\na&--&\cmark \\
		Canis&$2\,183$&4&2&\na&\na&\cmark&\cmark \\
		Lepus&$2\,183$&4&3&\na&\na&\cmark&\cmark \\
		Martes&$2\,183$&4&2&\na&\na&\cmark&\cmark \\
		Mammals&$2\,183$&4&7&\na&\na&\cmark&\cmark \\
		Octet&$82$&1&10&\na&\cmark&\cmark&\cmark \\
		\midrule
		\multicolumn{2}{l}{\textbf{Accuracy}} & & & $0.56$ & $0.82$ & $0.47$ & $0.88$ \\
		\bottomrule& & 
	\end{tabular}
	\caption{Comparison of \ltr, \ergo, \origo and \ourmethod on eleven multivariate data sets. We write {\na} whenever a method is not applicable on the pair.}
	\label{tab:mv_comparison}
\end{table}
	
\section{Conclusion}\label{sec:conc}

We considered the problem of inferring the causal direction from the joint distribution of two univariate or multivariate random variables $X$ and $Y$ consisting of single-, or mixed-type data. We point out weaknesses of known causal indicators and propose the Normalized Causal Indicator for mixed-type data and data with highly unbalanced domains. Further, we propose a practical encoding based on classification and regression trees to instantiate these causal indicators and provide a fast greedy heuristic to compute good solutions. 

In the experiments we evaluate the advantages of the NCI and the common indicator and give advice on when to use them. On real world benchmark data, we are on par with the state of the art for univariate continuous data and beat the state of the art on multivariate data with a wide margin.

For future work, we aim to investigate in the application of \ourmethod for causal discovery, meaning that we would like to infer causal networks. In addition, we only selected a subset of possible refinements to exploit dependencies from candidates. This choice could be expanded by considering more complex functions, finding combinations of categories for splitting. However, unless specific care is taken many of such extensions will likely have repercussions on the runtime of our algorithm, which is why besides being out of scope here, we leave this for future work.
	
%	%% only in camera ready version, or if we have space
	\section*{Acknowledgements}
	The authors wish to thank Kailash Budhathoki for insightful discussions. 
	Alexander Marx is supported by the International Max Planck Research School for Computer Science (IMPRS-CS). 
	Both authors are supported by the Cluster of Excellence ``Multimodal Computing and Interaction'' within the Excellence Initiative of the German Federal Government.
	
	% Bibliography
	\balance
	\bibliographystyle{abbrv} %
	\bibliography{abbrev,bib-jilles,bib-paper,bib-alex}%
	
\end{document}